# Improvement of a Prediction Model for Heart Failure Survival through Explainable Artificial Intelligence

Pedro A. Moreno-Sanchez [a,*]

[a] *School of Health Care and Social Work, Seinäjoki University of Applied Sciences, Seinäjoki, 60100 Finland*

**\* Corresponding author email address**: pedro.morenosanchez@seamk.fi

**Abstract**

Cardiovascular diseases and their associated disorder of heart failure are one of the major death causes globally, being a priority for doctors to detect and predict its onset and medical consequences. Artificial Intelligence (AI) allows doctors to discover clinical indicators and enhance their diagnosis and treatments. Specifically, "explainable AI" offers tools to improve the clinical prediction models that experience poor interpretability of their results. This work presents an explainability analysis and evaluation of a prediction model for heart failure survival by using a dataset that comprises 299 patients who suffered heart failure. The model employs a data workflow pipeline able to select the best ensemble tree algorithm as well as the best feature selection technique. Moreover, different post-hoc techniques have been used for the explainability analysis of the model. The paper's main contribution is an explainability-driven approach to select the best prediction model for HF survival based on an accuracy-explainability balance. Therefore, the most balanced explainable prediction model implements an Extra Trees classifier over 5 selected features (follow-up time, serum creatinine, ejection fraction, age and diabetes) out of 12, achieving a balanced-accuracy of 85.1% and 79.5% with cross-validation and new unseen data respectively. The follow-up time is the most influencing feature followed by serum-creatinine and ejection-fraction. The explainable prediction model for HF survival presented in this paper would improve a further adoption of clinical prediction models by providing doctors with intuitions to better understand the reasoning of, usually, "black-box" AI clinical solutions, and make more reasonable and data-driven decisions.

**Keywords**: Explainable Artificial Intelligence, medical XAI, Heart Failure, Clinical Prediction Models, Ensemble Trees

**Abbreviations**
- CVD: Cardiovascular Diseases
- HF: Heart Failure
- ML: Machine Learning
- AI: Artificial Intelligence
- XAI: eXplainable Artificial Intelligence
- EHR: Electronic Health Record
- PDP: Partial Dependence Plot
- SHAP: SHapley Additive exPlanations

# 1 Introduction

Cardiovascular diseases (CVD) are the global leading cause of death and disability with 17 million dead people approximately per year (31 % of the total deaths globally). In this decade (2020-2030), an increase from 31,5% to 32.5% will result into 3.7 million additional deaths worldwide ("Cardiovascular diseases (CVDs)," n.d.). In the US, the direct and indirect medical costs are expected to triple by 2030 respectively from $273 billion to $818 billion, and $172 billion to $276 billion. Therefore, it is crucial to develop preventive strategies to reduce CVD progression as well as minimizing the associated costs.

CVD term involves different disorders of the heart and circulatory system manifested in different pathologies as stroke, heart failure, or coronary heart disease. Heart Failure (HF) accounts for a large portion of the CVD morbidity and mortality, as well as a large portion of related healthcare expenses (Virani et al., 2020). HF appears when the heart is unable to pump correctly the blood to the rest of the body, and is accompanied by symptoms like shortness of breath or weakness (Ponikowski et al., 2016). HF is usually a consequence of other chronic diseases like diabetes or hypertension, as well as other patient's conditions like obesity, drug abuse or smoking (Virani et al., 2020). HF affects at least 26 million people globally and presents a high mortality rate (about 50% of HF patients will die within 5 years) (Savarese and Lund, 2017; Shameer et al., 2017). Due to the essential importance of the heart for a person's life, the HF onset prediction, as well as its consequences (e.g. mortality) has become a priority for doctors and healthcare providers not only for its patient's health implications but also because of the increase of resources associated to patient follow up (economic, humans, etc.). However, despite this urgent need, the clinical practice has failed so far to achieve high accuracy in these tasks (Chapman et al., 2019). Moreover, healing actions to HF patients tend to remain relatively minimal due to the difficult detection of the characteristics of these patients (Buchan et al., 2019).

As a consequence, modelling survival patients with HF remains currently challenging concerning the early identification of clinical factors associated with its mortality with a high classification accuracy (Kaddour, 2021). Nowadays, angiography is the most precise but costly solution to predict CVD, however, its high cost makes the access difficult for low-income families (Arabasadi et al., 2017). In this context, the increasing availability of electronic data implies an opportunity to democratize access to prediction models for HF survival. Machine Learning (ML) and Artificial Intelligence (AI), in the field of healthcare, has become a promising tool supporting clinicians to detect disease patterns, to predict risk situation for patients, and capture clinical knowledge from a large quantity of data. Computer-aided diagnosis systems, through ML algorithms implementation, offer a diagnosis of complex health issues with good accuracy and efficiency (Baby and Vital, 2015; Lakshmi et al., 2014). Therefore, ML is conceived as a vehicle to provide healthcare professionals with appropriate solutions to discover latent correlations between HF survival and clinical indicators enabling early detection of those patients at risk.

Nevertheless, if the decisions made by computer-aided diagnosis systems affect the patient's life, their use in clinical routine is not straightforward. In the healthcare domain, clinicians require far more information from the prediction models than a simple binary decision, thus, providing explanations that support the ML models' outputs is a cornerstone to ensure their adoption. The field of eXplainable Artificial Intelligence (XAI) arises to cope with this requirement, as it is defined as: "Given an audience, an explainable Artificial Intelligence is one that produces details or reasons to make its functioning clear or easy to understand" (Barredo Arrieta et al., 2020). In the medical context, the lack of explainability of certain prediction models must be addressed since clinicians find challenging to trust complex ML methods due to their high technical knowledge requirements (Carvalho et al., 2019). Thus, XAI would allow healthcare experts to make reasonable and data-driven decisions to provide more personalized and trustworthy treatments and diagnoses (Stiglic et al., 2020). However, XAI is not a "one-size-fits-all" solution because an inherent tension between accuracy and explainability appears depending on ML models employed, being usually the best-performing models those more complex and less interpretable (e.g. ensemble trees or neural networks) and vice versa.

This paper aims at describing the development of a prediction model for HF survival with a justified balance between accuracy and explainability. In addition, by applying different explainability post-hoc techniques to the model, the influence of the different clinical indicators on the prediction results are analyzed. To develop the explainable prediction model, an automated data pipeline is used to select different model's parameters like ensemble tree classifier or features selected that indicates the best classification performance.

The remainder of this paper is the following: Section 2 indicates the related works identified that build an HF survival prediction model with the same dataset used in this paper. Section 3 refers to the dataset, the different ML algorithms, feature selection methods and metrics employed in this work. Section 4 presents the pipeline employed to build the predictive model, the evaluation results in terms of classification and explainability, and the analysis of the features' importance for global and local explainability purposes. In section 5, the results obtained are discussed. Finally, Section 6 includes the conclusions drawn from the work.

## 2  Related Works

Within the healthcare domain, professionals' demand for tools that make AI more accessible is increasing over time since AI solutions usually requires expert knowledge about ML algorithms (Olson et al., 2016). This need is especially important in precision medicine where disease diagnosis requires interpretable and transparent information (Tjoa and Guan, 2020). Since more than a decade ago, XAI solutions aimed at giving healthcare professionals prediction models' global explanations have been used. In particular, transparent models like logistic and linear regression, naïve Bayes, decision tree or k-nearest neighbours were employed in different clinical fields, namely: urology (Otunaiya and Muhammad, 2019; Zhang et al., 2019), cardiology (Feeny et al., 2019), toxicology (Zhang et al., 2019, 2018), endocrinology (Sossi Alaoui et al., 2020), neurology (Zhang and Ma, 2019), psychiatry (Guimarães et al., 2019; Obeid et al., 2019), occupational diseases (Di Noia et al., 2020), knee osteoarthritis (Jamshidi et al., 2019), breast cancer (Aro et al., 2019), prostate cancer (Seker et al., 2000), severity of Alzheimer's disease (Bucholc et al., 2019), diabetes (Karun et al., 2019) and mortality rates of CVDs (e.g. myocardial infarction or perinatal stroke) (Prabhakararao and Dandapat, 2019) (Gao et al., 2017). Other explainability model-agnostic solutions like SHAP (SHapley Additive exPlanations) or MUSE (Model Understanding through Subspace Explanations) have been applied to complex AI solutions based on deep learning to diagnose depression (Lakkaraju et al., 2019), prevent hypoxemia during surgery (Lundberg et al., 2018), or image detection of acute intracranial hemorrhage (Lee et al., 2019).

HF outcome prediction is critical to apply accurately available therapeutic options, ranging from pharmacologic to highly invasive mechanical ventricular assistance and cardiac transplantation (Guo et al., 2020). ML techniques can be useful to predict risk at an early stage by using the variables derived from the complex and diverse EHR data of patients. Various accurate methods, like ADHERE model (Fonarow et al., 2005) and the Seattle Heart Failure Model (Levy et al., 2006), were developed during the last decade to estimate the risk of death for patients with HF, although they were unintuitive and rely on extensive medical records, making them hard to apply in a clinical setting (Wilstup and Cave, 2021). Other studies have been developed for classifying CVD diseases and to predict accurately abnormalities in the heart or its functioning (Ravish et al., 2014; Shah et al., 2020; Zhang and Han, 2017). Multiple varieties of ML algorithms have been employed for this CVD prediction models like for instance: Support Vector Machines, Logistic Regression, Artificial Neural Networks, Random Forest, Decision Tree, Ensemble Learning approaches, Deep Neural Networks, Fuzzy experts system, or K-nearest Networks (Ishaq et al., 2021). However, modelling survival heart failure is still uncovered in terms of driving factors identification since models developed presented limited interpretability of their prediction variables (Smith et al., 2011) ("Global mortality variations in patients with heart failure: results from the International Congestive Heart Failure (INTER-CHF) prospective cohort study - The Lancet Global Health," n.d.). Another issue found in the literature is the lack of consensus about HF indicators' relevance since studies employ different datasets which affect the models' reliability to be deployed in clinical routine (Alba et al., 2013; Guo et al., 2020). As a consequence, partial approaches tackle the model´s effectiveness through cohorts with specific types of patients (e.g. elderly o diabetic) [14] (Dauriz et al., 2017), although their models developed have not achieved an optimal performance (Segar et al., 2019) (Son et al., 2019).

Therefore, to proceed with an objective comparison to other HF prediction models, homogeneity concerning the data set must be maintained. The dataset released by Ahmad et al. (Ahmad et al., 2017) in UCI public repository (Dua and Graff, 2017) allows building a benchmarking of prediction models with other authors' work. Table 1, shows the most recent work that employs Ahmad's dataset to build a prediction model for HF survival. The reviewed works are sorted in a descendant order of accuracy (Acc.), although other information per each work is expressed like Sensitivity (Sens.), Specificity (Spec.), f1-score(F1), and Precision (Prec.); the number of features (#Feat.) after applying feature selection; and the Machine Learning (ML) technique. Note that since the dataset is imbalanced in its target feature, some works consider also balanced accuracy in their metrics.

## 3  Material and Methods

### 3.1 Heart Failure Survival dataset

The dataset employed in this paper, as mentioned before, was released by Ahmad et al. and contains the medical records of 299 patients (194 men and 105 women) who suffered an HF episode. The data set was collected from April to December 2015 at the Faisalabad Institute of Cardiology and at the Allied Hospital in Faisalabad (Punjab, Pakistan). 7 numerical and 5 nominal features composed the dataset along with one binary target feature ("death event"). This dataset presents an imbalance concerning its target feature since 203 out of the 299 instances belongs to patients who survived to HF ("death event"=0), and the rest of 96 instances represents patients deceased ("death event"=1). All instances of the dataset are entirely complete with no missing values in any of their features. The dataset description is depicted in Table 2

**Table 1.** Classification results of related works and ML classifiers (best ones in italic). (Unk: unknown; * Studies that perform the best classifier over unseen new data)

| Author | Acc. (Bal Acc) | Sens. | Spec. | Prec. | F1 | #Feat | ML Classifer |
|---|---|---|---|---|---|---|---|
| (Rahayu et al., 2020) | 0.83 | - | - | - | - | - | Random Forest, Decision Tree, K-Nearest Neighbour, *Support Vector Machine*, Artificial Neural Network, Naïve Bayes |
| (Kumar et al., 2021) | 0.96 | 0.93 | - | 0.95 | 0.94 | 5 | *Random Forest*, XGBoost, Decision Tree, Logistic Regression, Support Vector Machine, K-Nearest Neighbour, Gradient Boosting, Stochastic Gradient Descent, Gaussian Naïve Bayes |
| (*) (Ishaq et al., 2021) | 0.88 | 0.89 | - | 0.89 | 0.89 | - | *Random Forest*, XGBoost, Decision Tree, AdaBoost, Extra Trees, Logistic Regression, Support Vector Machine, Gradient Boosting, Stochastic Gradient Descent, Gaussian Naïve Bayes |
| (*) (Kaddour, 2021) | 0.90 (0.91) | 0.93 | 0.90 | - | - | 4 | FeedForward Neural Network, *Deep Neural Network* |
| (Kucukakcali et al., 2020) | 0.87 (0.82) | 0.69 | 0.95 | - | 0.77 | - | *Associative Classification* |
| (Wilstup and Cave, 2021) | 0.82 | - | - | - | - | 3 | *Cox models plus symbolic regression* |
| (Khan et al., 2021) | 0.81 | 0.82 | 0.74 | - | - | 5 | Support Vector Machine (Kernel Linear, Radial Basis Funcion, *Cubic* and Quadtratic) |
| (Taj et al., 2021) | 0.72 | - | - | - | - | 7 | *Fuzzy Preti nets plus Rough Set Theory* |
| (Gürfidan and Ersoy, 2021) | 0.83 | - | - | - | - | - | *Support Vector Machines*, Logistic Regression, Decision Tree, K-Nearest Neighbour, Linear Discriminant Analysis, Gaussian Naïve Bayes |
| (Chicco and Jurman, 2020) | 0.84 | 0.78 | 0.86 | - | 0.72 | 3 | *Random Forest,* Gradient Boosting, Support Vector Machine with radial kernel |

### 3.2 Ensemble Trees algorithms

Ensemble trees techniques, by weighting and combining various models generated from a base decision tree, usually offer reasonably good accuracy in classification tasks and are usually employed in different research fields (health, economy, biology, etc) (Sagi and Rokach, 2020). These ensemble methods not only outperform the weak base classifier but also allow mitigating challenges as class imbalance or the curse of dimensionality (Sagi and Rokach, 2020).

However, due to the lack of explainability capabilities, ensemble trees might be avoided by professionals who needs to interpret the predictions. As a consequence, post-hoc explainability techniques are needed to interpret the black-box behaviour of ensemble trees. The different ensemble trees algorithm employed in this work are described in

Table *3*.

### 3.3 Explainability techniques for ML

Regarding explainability, decision tree are denoted as a "transparent" model because of its graphical structure and decomposability that provide a fully interpretable functionality making them appropriate for those domains where ML models' outputs are required to be understood (eg. healthcare).

As mentioned before, ensemble trees require support from post-hoc explainability techniques since their classification decision is based on the combination of multiple decision trees' results. Post-hoc explainability techniques offer understandable information about how an already developed model produces its predictions through the employment of common methods that humans use to explain systems like visual, local or feature relevance explanations, which are used in this research (Barredo Arrieta et al., 2020). As following, different post-hoc techniques in this research are described.

**Table 2.** Dataset's features description

| Id | Feature | Range (mean ± std) / binary values (number of instances per class) |
|---|---|---|
| 1 | Age (years) | 40-95 (60.83 ± 11.89) |
| 2 | Anaemia (boolean) | 0 (170) or 1(129) |
| 3 | High Blood Pressure (boolean) | 0 (194) or 1 (105) |
| 4 | Creatinine phosphokinase-CPK(mcg/L) | 23-7861 (581.83 ±970.29) |
| 5 | Diabetes (boolean) | 0 (174) or 1 (125) |
| 6 | Ejection fraction (percentaje) | 14-80 (38.08 ± 11.83) |
| 7 | Sex (boolean) | 0 (194) or 1 (105) |
| 8 | Platelets (kiloplatelets/mL) | 25100-850000 (263358.03 ± 97804.23) |
| 9 | Serum creatinine (mg/dL) | 0.50-9.40 (1.39 ± 1.03) |
| 10 | Serum sodium (mEq/L) | 113-148 (136.62 ± 4.41) |
| 11 | Smoking (boolean) | 0 (203) or 1(96) |
| 12 | Time-Follow up period (days) | 4-285 (130.26 ±77.61) |
| 13 | [Target] Death event (boolean) | 0 (203) or 1(96) |

**Table 3.** Ensemble Trees algorithms employed

| Ensemble Trees algorithm | Description |
|---|---|
| Decision Tree (base classifier) | Decision trees depict a flow chart that employs branching methods to represent possible outcomes of a decision based on instance's different values. The classification rules can be derived from the path established from the root to the leaf node that generates the decision (Dessai, 2013). |
| Random Forests | Due to a fairly good predictive performance as well as the capability to deal with datasets of different sizes, *Random Forest* is one of the ensemble trees methods most widely used. Random Forest employs, to train its base classifier (decision tree), the bagging method that selects randomly a group of features at splitting in its nodes (Sagi and Rokach, 2020) |
| Extreme Randomized Trees (Extra Trees) | *Extra Trees* aim to improve accuracy of the tree-based bagging classifiers by selecting random cut-points in the nodes splitting as well as by using the total learning samples by all classifier base trees (Geurts et al., 2006). |
| Adaptive Boosting (AdaBoost) | AdaBoost focuses its training process on misclassified instances that receive modified weights over successive iterations. In addition, the base classifiers also receive certain weights according to their performance that influences the classifying output of a new instance (Sagi and Rokach, 2020) |
| Gradient Boosting | Gradient boosting trains their base classifier over the residual errors from the precedent classifiers, hence, reducing the classification error. The overall classification result is obtained through a weighted average of all base classifiers' results (Friedman, 2002) |
| eXtreme Gradient Boosting (XGBoost) | XGBoost applies several optimizations and regularization processes to the gradient boosting algorithm in order to increase the speed and performance as well as make the algorithm simpler and more generative (Sagi and Rokach, 2020). |
| Ensemble voting classifier (Max Voting) | It performs by aggregating the prediction of multiple different classifiers. In the "max-voting" mode, only the class with the highest votes is included as the final prediction. |

*3.3.1   Feature Permutation Importance*

Permutation feature importance as a post-hoc technique, based on local explanations, measures the increase in the prediction error of the model after permuting a specific feature's values (Fisher et al., 2019). This model-agnostic technique (non-dependent on the ML technique to explain) indicates a feature as important if the error increases by shuffling the feature's values a specific number of times. Vice versa, if the error does not change by shuffling the feature's values, the feature is "unimportant".

*3.3.2 Partial Dependence Plot*

Partial Dependence Plot (PDP) is a visual explanation post-hoc explainability technique that shows the marginal effect of a given feature on the predicted outcome (Friedman, 2000). The marginal effect concept indicates how a dependent variable changes when a specific independent variable changes its values, meanwhile other covariates remain constant Therefore, PDP can be used as a model-agnostic method for global explainability to determine the average effect of a feature over a range of different observed values. For classification tasks, such as the one performed in this study, the PDP displays the probability (average and confidence interval) for a certain class as a function of the feature value. PDP also offer a multivariate option, where for instance the marginal effect of two features can be analysed over the output probability.

*3.3.3 Shapley Values*

The SHapley Additive exPlanations (SHAP) technique is a model-agnostic method that combines explanations by example with feature relevance. The technique computes an additive feature importance score for each individual prediction with local accuracy and consistency (Lundberg and Lee, 2017). SHAP computes the contribution of each feature to the predicted outcome/class by applying coalitional game theory (Lundberg et al., 2018). In classification tasks, the SHAP technique computes a signed importance score that indicates the weight of a feature towards the predicted outcome as well as its direction where positive values increase the probability of class 1 and negative ones decrease such probability.

*3.3.4 Feature Selection*

Feature selection cannot be considered as a specific explainability technique; however, it can enhance model explainability since when it is performed during the data preparation phase those unimportant features that bring non-relevant information to the classification are removed. Feature selection also allows decreasing overfitting in models' prediction, and even reducing computing time. Moreover, searching for a relevant features subset involves finding those features that are highly correlated with the target feature, but uncorrelated with each other (Shilaskar and Ghatol, 2013).

Generally, there are three types of feature selection methods: filters methods where intrinsic properties of data justify the inclusion of an attribute or a subset of attributes; wrappers methods similar to filters except a classification algorithm is utilized; and embedded methods that combine filter and wrapper to achieve a better classification performance. Concerning filter methods, different techniques are applied depending on the data type of the features and the target variable (Kuhn and Johnson, 2013), for instance, ANOVA correlation coefficients are used in the case of numerical input and categorical output, and Chi-Squared test when both categorical input and output. Mutual information is another filter method applied when the output variable is categorical but do not depend on the input data type. As regards wrapper methods, one of the most frequently employed is Recursive Feature Elimination (RFE) that use an estimator, like logistic regression, to reduce recursively the features in a dataset by discarding those features with the smallest weights during recursive iterations. These four methods have been considered in this research.

*3.4 Performance metrics*

Different metrics are considered in this paper to evaluate the classification performance of the prediction model as well as its explainability. Table 3 summarize the formulas of these metrics.

Accuracy measures the rate of true predictions in all classifications made with a dataset, and it is a recommendable metric when dealing with balanced datasets. However, the dataset employed in this work is not balanced in its target feature, thus, the accuracy metric can give a wrong idea about the model's classification performance. Thus, the balanced accuracy (BAcc) gives a better insight since it accounts for the imbalance in classes. The rest of the metrics considered are especially useful when evaluating a classification model within the healthcare domain where false positive and false negative are important (Zhang et al., 2019).

On the other hand, considering ensemble trees as the classifiers employed, the explainability metrics used are Interpretability, Fidelity and Fidelity-Interpretability Ration proposed by Tagaris et.al (Tagaris and Stafylopatis, 2020). The model's interpretability, I (model), can be defined as the percentage of those masked features that do not bring information to the final classification result and the total number of features of the dataset. The model's fidelity, F(model), is calculated through the relation of a specific classification metric results (eg. accuracy) of the full-interpretable model and its un-interpretable counterpart. A model's fidelity shows the percentage of its initial performance that it managed to retain by becoming interpretable. Finally, the model's Fidelity-to-Interpretability Ratio (FIR) shows how much of the model's interpretability is sacrificed for performance. A balanced ratio of 0.5 is the optimal score.

**Table 4.** Classification and explainability metrics formulas (TN as true negative, FN as false negative, FP as false positive, and TP as true positive).

| Ensemble Trees algorithm | Description |
|---|---|
| *Accuracy* | $\frac{(TP+TN)}{(TP+TN+FP+FN)}$ (1) |
| *Sensitivity/Recall* | $\frac{TP}{(TP+FN)}$ (2) |
| *Specificity* | $\frac{TN}{(TN+FP)}$ (3) |
| *Balanced accuracy* | $\frac{Sensivity + Specificity}{2}$ (4) |
| *Precision* | $\frac{TP}{(TP+FP)}$ (5) |
| *F1-Score* | $2*\frac{Precision*Sensivity}{Precision+Sensivity}$ (6) |
| *Interpretability (I)* | $\frac{masked\ features}{total\ input\ features}$ (7) |
| *Fidelity (F)* | $\frac{P(model)}{P(baseline)}$ (8) |
| *Fidelity-Interpretability Ratio* | $\frac{F}{F+I}$ (9) |

*3.5 Data workflow pipeline*

To develop the explainable prediction model for HF survival, the automated data workflow pipeline published in (Moreno-Sanchez, 2021) and named SCI-XAI was used. This pipeline, shown in Fig. 1 implements the automation of several phases of CRISP-DM methodology (Wirth and Hipp, n.d.), namely: data preparation, modelling and evaluation.
The SCI-XAI pipeline uses the GridSearchCV module of python scikit-learn package (Pedregosa et al., n.d.) that allows applying a brute force algorithm to find the best combination, in terms of classification performance, of ensemble trees classifier technique, the number of features selected, method of feature selection and data missing imputation method.

As a first step, the original dataset is divided into two sub-datasets, training and test, with 280 and 120 respectively. Thus, a split ratio of 70/30 is adopted with a stratification approach that ensures the same proportion of the target feature (Death_event) in both sets. This initial split is aimed at building the prediction model using exclusively the train set's instances, to evaluate next the model's performance with unseen new data stored in the test set. By doing so, any influence of the test set´s instances in feature selection and classifier training modules is avoided. Then, the data preparation phase comprises the modules of data missing, normalization in case of numerical features, encoding for nominal and ordinal features and finally the feature selection. As Fig 1 shows, this data preparation handles features depending on their type (numerical, nominal and ordinal). The modelling or training phase is carried out using a 5-fold cross-validation approach through the GridSearchCV to fit different types of ensemble trees classifiers on the training data. The best model selected is finally evaluated in terms of classification and explainability.

The source code of this work can be found in https://github.com/petmoreno/Heart_Failure_Predictor

# 4 Results

*4.1 Classification performance*

The results obtained by applying the different ensemble tree learning algorithms in the 5-fold cross-validation training module are depicted in Table 5. It must be noted that the results in the table denote, in the different metrics considered, the best performance for each classifier that intrinsically engages a group of features selected. To facilitate the comparison between the classifier´s performance to a baseline case when dealing with the whole feature dataset, Table 6 shows the pipeline's metric results without applying the features selection module. Finally, the results obtained with new unseen data (and feature selection module) are also shown in Table 7.

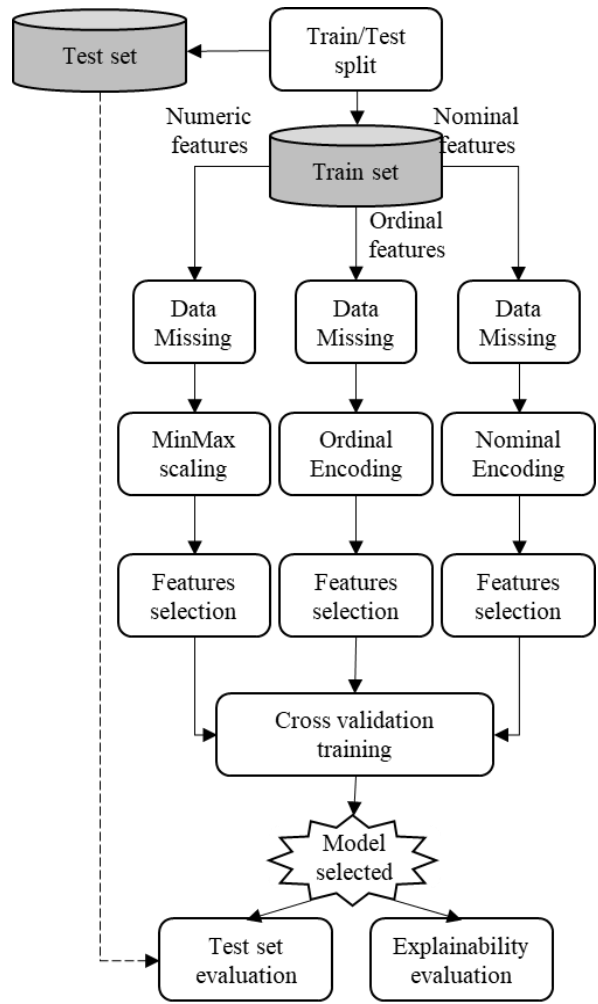

Fig. 1 SCI-XAI automated data workflow pipeline

Table 5. Classification results of the Training set (Cross-validatoin approach) with feature selection

| Classifier | Acc. | BAcc. | Sens. | Spec. | Prec. | F1 |
|---|---|---|---|---|---|---|
| Random Forests | 0.876 | 0.854 | 0.792 | 0.916 | 0.820 | 0.804 |
| Extra Trees | 0.871 | 0.851 | 0.793 | 0.908 | 0.807 | 0.799 |
| AdaBoost | 0.852 | 0.826 | 0.751 | 0.901 | 0.787 | 0.764 |
| Gradient Boosting | 0.847 | 0.830 | 0.780 | 0.880 | 0.760 | 0.766 |
| XGBoost | 0.871 | 0.850 | 0.792 | 0.908 | 0.806 | 0.797 |
| Max Voting | 0.866 | 0.839 | 0.763 | 0.915 | 0.816 | 0.788 |

Table 6. Classification results of the Training set (Cross-validation approach) without feature selection

| Classifier | Acc. | BAcc. | Sens. | Spec. | Prec. | F1 |
|---|---|---|---|---|---|---|
| Random Forests | 0.852 | 0.812 | 0.701 | 0.923 | 0.810 | 0.750 |
| Extra Trees | 0.818 | 0.753 | 0.568 | 0.937 | 0.806 | 0.665 |
| AdaBoost | 0.823 | 0.784 | 0.674 | 0.894 | 0.766 | 0.714 |
| Gradient Boosting | 0.838 | 0.803 | 0.705 | 0.901 | 0.771 | 0.735 |
| XGBoost | 0.847 | 0.822 | 0.749 | 0.895 | 0.775 | 0.757 |
| Max Voting | 0.838 | 0.790 | 0.657 | 0.923 | 0.804 | 0.720 |

Table 7. Classification results of Test set (new unseen data) with feature selection

| Classifier | Acc. | BAcc. | Sens. | Spec. | Prec. | F1 |
|---|---|---|---|---|---|---|
| Random Forests | 0.800 | 0.753 | 0.621 | 0.885 | 0.720 | 0.667 |
| Extra Trees | 0.844 | 0.795 | 0.655 | 0.934 | 0.826 | 0.731 |
| AdaBoost | 0.833 | 0.778 | 0.621 | 0.934 | 0.818 | 0.706 |
| Gradient Boosting | 0.800 | 0.735 | 0.552 | 0.918 | 0.762 | 0.640 |
| XGBoost | 0.789 | 0.718 | 0.517 | 0.918 | 0.750 | 0.612 |
| Max Voting | 0.811 | 0.752 | 0.586 | 0.918 | 0.773 | 0.667 |

## 4.2 Feature selection

The SCI-XAI pipeline not only extracts the best performance for each classifier but also the optimal number of features that contribute to such performance. **Error! Not a valid bookmark self-reference.** and Table 9 shows for numerical and nominal features respectively the number of features selected, their name as well as the technique employed, ie. ANOVA, chi-squared, Mutual information (mut-inf), or recursive feature elimination (RFE).

Table 8. Numerical features selected (# feats: number of features)

| Classifier | # feats | Features (select method) |
|---|---|---|
| Random Forests | 4 | age', 'ejection_fraction', 'serum_creatinine', 'time' (ANOVA) |
| Extra Trees | 4 | 'age', 'ejection_fraction', 'serum_creatinine', 'time' (ANOVA) |
| AdaBoost | 5 | age', 'ejection_fraction', 'serum_creatinine', 'serum_sodium', 'time'(ANOVA) |
| Gradient Boosting | 4 | creatinine_phosphokinase', 'ejection_fraction', 'serum_creatinine', 'time'(mut-inf) |
| XGBoost | 4 | creatinine_phosphokinase', 'ejection_fraction', 'serum_creatinine', 'time'(mut-inf) |
| Max Voting | 4 | creatinine_phosphokinase', 'ejection_fraction', 'serum_creatinine', 'time'(mut-inf) |

Table 9. Nominal features selected (# feats: number of features)

| Classifier | # feats | Features (select method) |
|---|---|---|
| Random Forests | 2 | 'diabetes', 'high_blood_pressure' (mut-inf) |
| Extra Trees | 1 | 'diabetes' (mut-inf) |
| AdaBoost | 1 | 'anaemia' (chi-squared) |
| Gradient Boosting | 2 | 'diabetes', 'high_blood_pressure' (mut-inf) |
| XGBoost | 5 | 'anaemia', 'diabetes', 'high_blood_pressure', 'sex', 'smoking' (RFE) |
| Max Voting | 1 | 'diabetes' (mut-inf) |

## 4.3 Explainability performance

When detecting the best combination of relevant features for each classifier, the evaluation of explainability can be performed. In order to calculate the Fidelity a pair of things are required to be decided: first, the balanced accuracy is chosen as the classification metric for the formula; and second, the decision tree performance is being used as the numerator since it is considered a fully interpretable model. The results of these metrics are shown in Table 10.

Table 10. Explainability metrics results

| Classifier | Interpretability | Fidelity | FIR |
|---|---|---|---|
| Random Forests | 0.50 | 0.94 | 0.65 |
| Extra Trees | 0.58 | 0.89 | 0.61 |
| AdaBoost | 0.50 | 0.91 | 0.65 |
| Gradient Boosting | 0.50 | 0.97 | 0.66 |
| XGBoost | 0.25 | 0.99 | 0.80 |
| Max Voting | 0.58 | 0.95 | 0.62 |

Considering FIR as the metric that gives a balanced measure between interpretability and fidelity, which are respectively related to explainability and accuracy, and being the value FIR=0.5 the optimal point, the HF survival prediction model built by Extra Trees with an FIR value of 0.61 can be denoted as the most balanced among those evaluated. Therefore, the prediction model built with Extra Trees and its group of selected features is used for conducting the explainability analysis.

*4.4    Explainability analysis of the prediction model*

As regards the explainability assessment of the ensemble trees algorithms consider in this work, the HF survival prediction model built with the Extra Trees classifier is the most balanced model in terms of explainability and accuracy. Therefore, in this subsection, the relevance of the following features 'age', 'time', 'diabetes' 'ejection_fraction' and 'serum_creatinine', are analysed to show their influence in the prediction task. As following, different post-hoc explainability techniques are implemented on the selected prediction model.

*4.4.1    Implicit feature relevance*

The implementation of Extra Trees in sci-kit learn allows extracting for each feature the Gini importance known as "mean decrease impurity". Fig. 2 shows the importance ranking of features selected based on Gini where weights and its standard deviation for the features are: 'time' ($0.457 \pm 0.089$), 'serum_creatinine' ($0.200 \pm 0.055$), 'ejection_fraction' ($0.190 \pm 0.044$), 'age' ($0.121 \pm 0.040$) and 'diabetes' ($0.032 \pm 0.020$). Moreover, Fig. 3 shows the features' importance and their attribution to the final score (negative or positive) in the case of a true negative (y=0, the patient is alive) and a true positive classification (y=1, the patient is deceased).

Either for global explainability or a particular instance prediction, the time follow up ('time') feature seems to be the most relevant. However, the rest of the features vary their importance depending on the situation. It must be noted that BIAS term refers to the proportion of instances where y=0 or y=1 for the whole dataset.

*4.4.2    Feature Permutation Importance*

As Fig. 4 shows, when permuting the values of the feature 'time', the prediction error suffers the biggest increase than with other features. Thus, feature permutation technique denotes 'time' (weight: 0.260, std: 0.0153) as the most relevant feature, followed in importance descending order by 'ejection_fraction' (weight: 0.097, std: 0.008), 'serum_creatinine' (weight: 0.068, std: 0.018), 'age' (weight: 0.029, std: 0.009) and 'diabetes' (weight: 0.022, std: 0.009).

*4.4.3    Partial Dependence Plot*

This post-hoc visual explanation provides the relation trend (marginal effect) or the influence direction between the target feature and the distribution values of the rest of the features: 'time' (Fig. 5), 'ejection_fraction' (Fig. 6), 'serum_creatinine' (Fig. 7), 'age' (Fig. 8) and 'diabetes' (Fig. 9). By exploring the PDP curve, certain values where the marginal effect curve change can be identified allowing the expert professionals to establish certain thresholds, intervals or trigger values that affect the prediction probability. In addition, negatives values of the curve manifest an inverse influence on the target outcome and vice versa.

Therefore, 'age', 'diabetes' and 'serum_creatinine' present a very small influence on the outcome probability, except in the interval between 0-2 for 'serum_creatinine' that shows a negative effect, meanwhile 'age' would present a threshold at the value of 50 years that influences positively the likelihood of a death event. In the case of 'time' and 'ejection_fraction', the influence is below 0 for every value of their distributions and monotonic from the values of 180 days and 30% respectively. Regarding 'time', the effect shows a decreasing quadratic trend during the first 100 days, and a soft peak about 120 days approximately. A gentle negative slope is depicted for values concerning 'ejection_fraction' between 0 and 25%, which decreases until 30%.

PDP also allows analysing the marginal effect caused on the target by two features. Given a bigger influence of 'ejection_fraction', 'serum_creatinine' and 'time' than the other features ('age' and 'diabetes'), three 2-dimensional PDP are depicted, namely: 'time'-'ejection_fraction' (Fig. 10), 'time'-'serum_creatinine' (Fig. 11), and 'ejection_fraction'-'serum_creatinine' (Fig. 12). These 2D-PDP figures show of lower values of 'time' is perceived as relevant since they raise the probability higher than 0.6 at any value of 'ejection_fraction' and 'serum_creatinine'. Moreover, these latter features have a relevant influence when low values of 'ejection_fraction' and high 'serum_creatinine'.

*4.4.4    Shapley values*

By using the SHAP library (Lundberg et al., 2020, 2018), the Shapley values technique can be applied to analyse global and local explainability on a certain classifier, Extra Tree in our case. SHAP allows depicting the global influence for different prediction classes, ie. Class 0 (y=0) and class 1 (y=1). Thus, Fig. 13 shows each feature' importance through the average impact on the model.

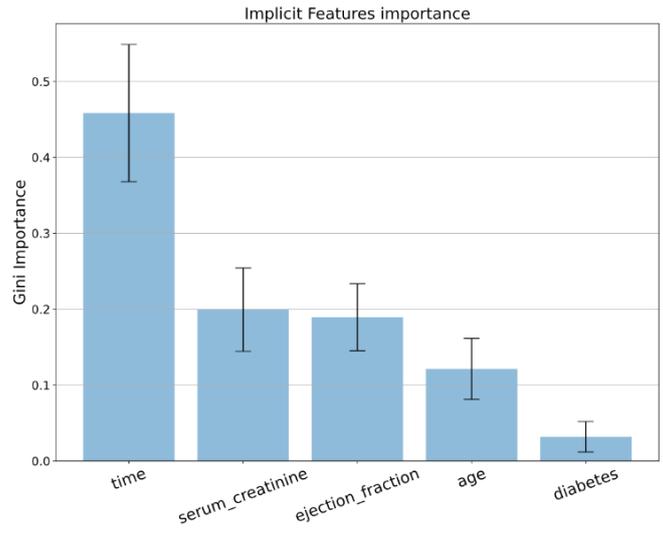

**Fig. 2**. Implicit Feature Importance for Global explainability

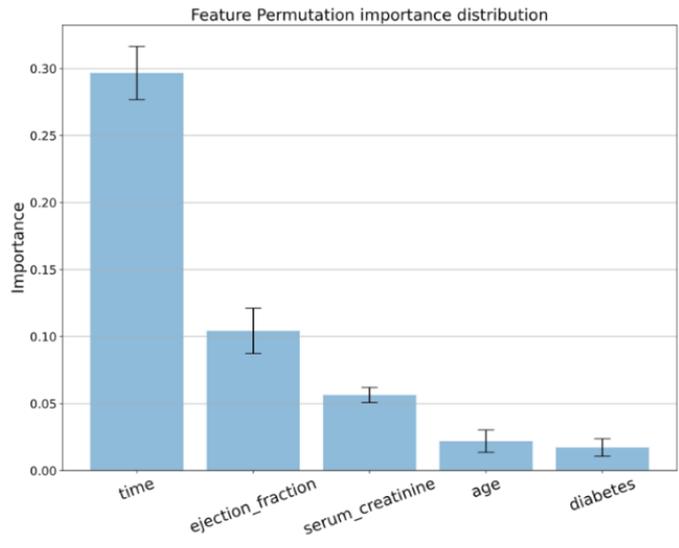

**Fig. 3**. Implicit Feature Importance for true negative (y=0) and true positive prediction (y=1).

**Fig. 4**. Feature permutation importance distribution.

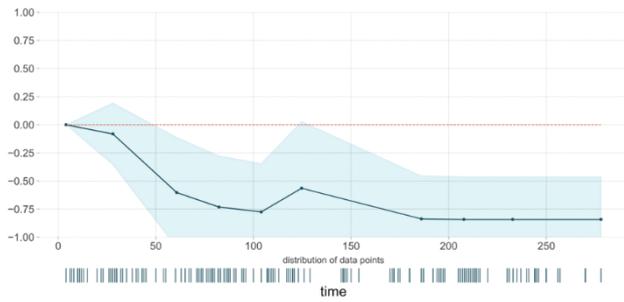

**Fig. 5**. PDP of feature time

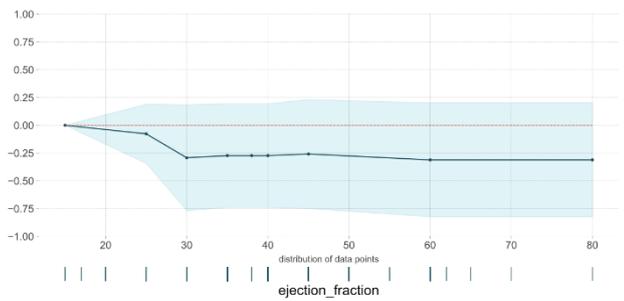

**Fig. 6**. PDP of feature ejection_fraction

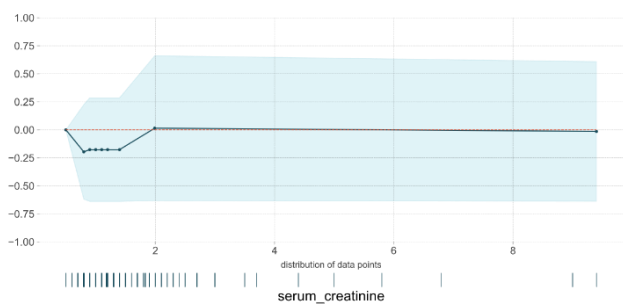

**Fig. 7**. PDP of serum_creatinine

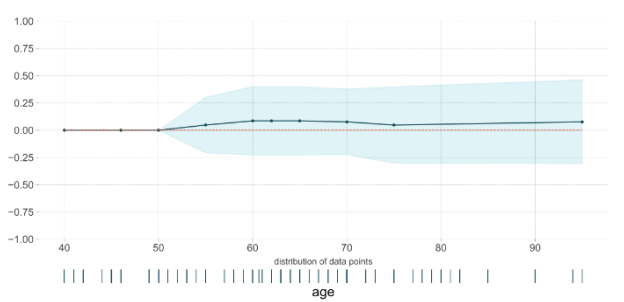

**Fig. 8**. PDP of feature age

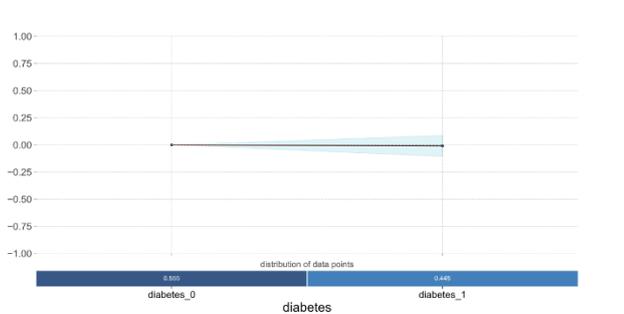

**Fig. 9**. PDP of feature diabetes

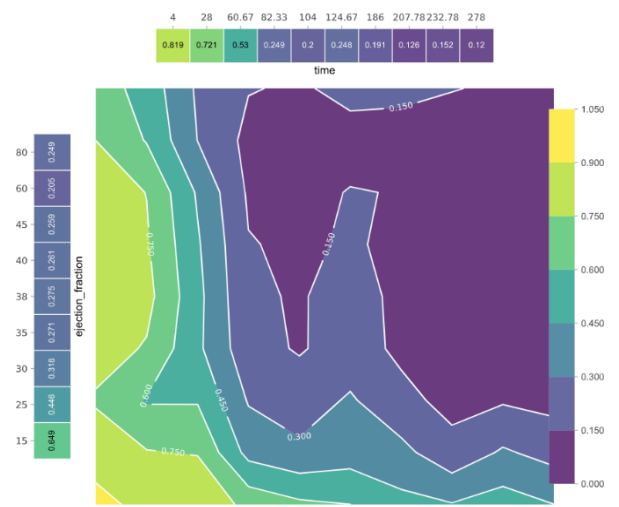

**Fig. 10**. 2D PDP of time and ejection_fraction

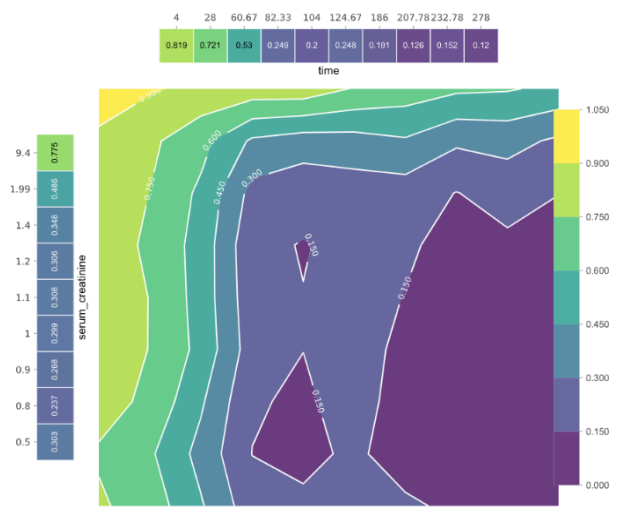

**Fig. 11**. 2D PDP of time and serum_creatinine

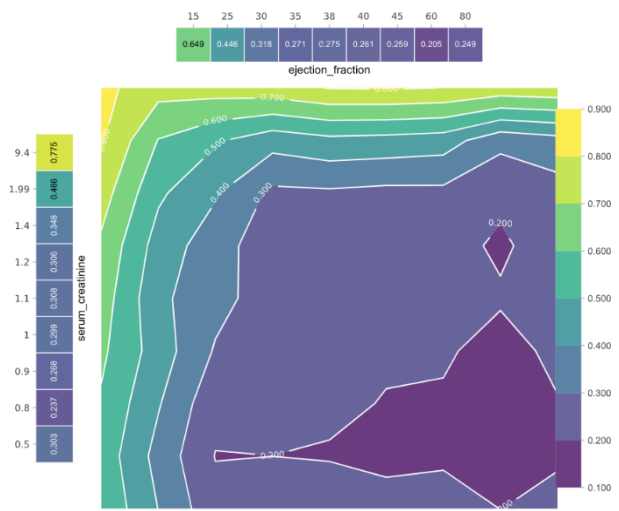

**Fig. 12**. 2D PDP of ejection_fraction and serum_creatinine

On the other hand, SHAP depicts, through its "waterfall plots", the attribution of each feature when dealing with local instance prediction not only specifying the feature's weight (length of the bar) but also its direction towards the final score (positive or negative sign). Thus, Fig. 14 and Fig. 15 shows the features' attributions when predicting, respectively, a true negative case (y=0, survival patient) and a true positive case (y=1, deceased patient). These two prediction examples are the same as referred in the implicit feature importance results.

## 5   Discussion

Due to the importance of CVD in the current global context of chronic diseases increase, the prediction of their outcomes like survival or disease onset by applying ML could become a priority tool for doctors to achieve early identification of those factors related to the disease's effects. Moreover, XAI might represent an advance to those prediction models by covering clinicians' understandability requirements on the decision made by the models. XAI might also contribute to widening the prediction models' adoption in clinical practice since the professionals are enabled to make more reasonable and data-driven decisions. With more explainable clinical prediction models, doctors could focus on controlling those underlying features or indicators, and trying to reverse a worsening in the condition of a patient who has suffered, in this case, an HF.

This paper aims at describing a prediction model for HF survival aimed to facilitate the early detection of indicators that can lead the patient to death. This prediction model has been developed pursuing not only high prediction accuracy but also analysing the explainability of its results. This research contributes to enlarge the works dedicated to HF survival prediction by using ML through a novelty perspective, to the best of our knowledge, that tackle the model's explainability as a relevant part of the model. By employing post-hoc explainability techniques, this work support "opening" the black-box paradigm of the ensemble trees classifiers employed in clinical prediction models.

This prediction model for HF survival has been developed through a data management pipeline, previously developed in other author's work, that automates the data preparation, modelling and evaluation phases defined by the CRISP-DM methodology (Wirth and Hipp, n.d.). By using this pipeline different parameters like the appropriate ensemble tree algorithm, relevant features selected, and data imputation techniques can be automatically inferred to find the optimal prediction model in terms of classification and interpretability as well as improving its efficiency and scalability. To ensure the model's robustness against new unseen data, the pipeline performs a double evaluation of the model's performance by splitting initially the dataset for training and testing purposes.

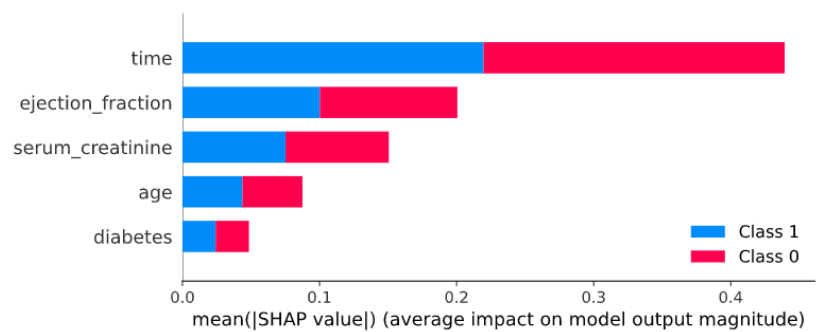

**Fig. 13**. Global explainability of HF survival prediction model by using SHAP.

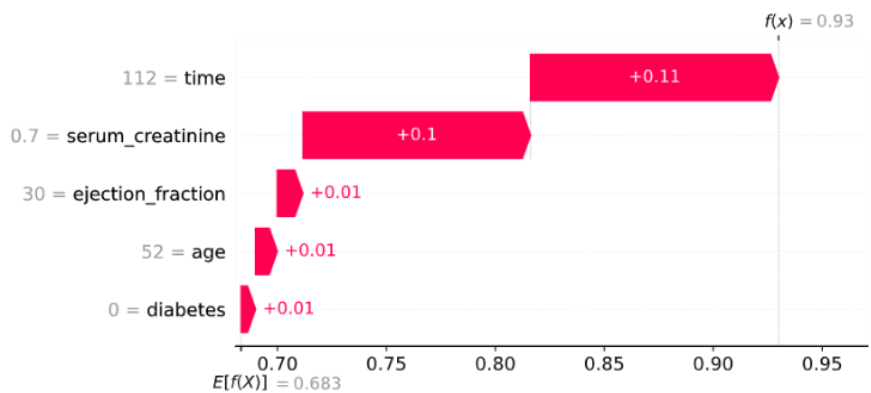

**Fig. 14**. Local explainability for True Negative case by using SHAP.

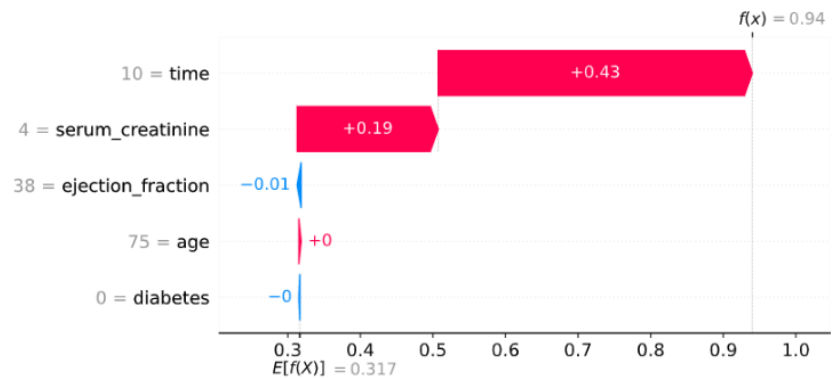

**Fig. 15**. Local explanability for True Positive case by using SHAP.

To detect the best prediction model, several ensemble tree methods, along with different combinations of features selected, have been considered due to the classification good performance of these algorithms (i.e. Random forest, Extra Trees, AdaBoost, gradient boosting, XGBoost, and max voting). To train and validate the model, a 5-fold cross-validation approach was used obtaining a classification performance with an accuracy of 84.7-87.6% or a balanced accuracy between 81-85.4% in all ensemble trees algorithms, being Random Forest the best with 87.6% and 85.4% respectively. It must be noted that the balanced accuracy is conveniently used because of a presence of an imbalance in the target feature (203 for y=0 and 96 for y=1). In addition, the pipeline was also used without the feature selection module, obtaining slightly worse results in the cross-validation that denotes the presence of non-relevant features that decrease the classification performance.

Furthermore, classification performance decreases generally when dealing with instances of the test set. In the case of test set evaluation, the Extra Tree is the best classifier with 84.5% and 79.5% of accuracy and balanced accuracy respectively. Therefore, the application of the SCI-XAI pipeline shows good results, placing this work in the top 4 of the related works in terms of accuracy classification. It must be highlighted that the test set involves 30% of the entire dataset which could emulate a deployment environment where the model deals with new unseen data. However, the model's performance in a real clinical environment might differ from the results because the medical records are not usually curated and might have a large number of features which could imply more complex patterns.

Moreover, these results demonstrate the capability of the pipeline to detect relevant features when building the prediction model by discarding around 50% (except for XGBoost with 25%) from the original number of features. The fewest features selected (5 out 12) is achieved when implementing the Extra Trees and Max Voting. To the best of our knowledge, no paper that has addressed an explainability evaluation of a prediction model for HF survival as it is proposed in this work. The Interpretability metric shows the relation of those non-relevant features to the original ones, meanwhile the Fidelity metric denotes the percentage that the model needs to "sacrifice" to become interpretable with its equivalent decision tree. By considering the FIR value of 0.5 as the optimal point in terms of an equilibrium between accuracy and explainability, the model that uses the Extra Trees is selected as the most balanced. Therefore, the best prediction model implements the Extra Trees algorithm and select the following feature as relevant: over a dataset with the following features: 'age', 'diabetes', 'ejection_fraction', 'serum_creatinine', and 'time'.

Concerning the explainability analysis of the prediction model by assessing the features' importance, the 'time' feature (follow-up time) seems to be the most relevant for all post-hoc techniques considered. This might be obvious since the longer the patient is being followed up the less probability to die. Thus, an additional study could arise by withdrawing the 'time' feature from the dataset. however, accuracy would decrease substantially justified by the importance gap between time and the rest of the features. Following a descending order of features' importance, 'serum_creatinine' and 'ejection_fraction' could be placed in a second-tier attending the global explainability results, although their importance weight seems to be interchanged depending on the post-hoc technique employed. In a third level, 'age' and 'diabetes' offer a small attribution to the global results, albeit, 'age' shows in most of the techniques far more important than diabetes. Particularly, PDP plots provide with an interesting opportunity to identify thresholds, intervals or values where a certain feature might increase or reduce the probability of the prediction, implying that doctors could indicate a treatment aimed to placed patient's features in those safe values where the outcome probability (mortality in this case) decreases. The utility of local explanations wants to be remarked since it could give the expert each feature's importance, which might differ from the global importance, for a certain specific patient. Explainability for true negative and true positive cases are shown in the results, where a remarkable difference in the attribution of 'serum_creatinine' and 'ejection_fraction' (especially in SHAP techniques) can be appreciated. With these local explainability results, it can be understood how the explainability techniques could contribute to the promotion of precision or personalized medicine.

This small heterogeneity in the results, notably in those less relevant features, manifests the need to count on different explainability techniques to reach a consensus on the features' importance. By doing so, doctors are able to focus on the specific indicators that could imply a change in the patient's health situation. Therefore, with the results described in this work, the added value of explainability to clinical prediction models is exhibited. Specifically, by adopting post-hoc explainability techniques as well as feature selection, a baseline prediction model that deals with all features of the original dataset is improved not only in terms of accuracy but also explainability. In addition, by offering a balance between these two latter terms, the prediction model for HF survival could imply a valuable tool for healthcare experts and increase its possibilities for being adopted in clinical routine.

# 6 Conclusions

The development and evaluation of an explainable prediction model for HF survival are presented in this work with the aim of showing the importance of considering explainability in early diagnosis clinical systems based on machine learning. The prediction model developed is improved by adopting a balanced approach between the model's classification performance and its explainability which could make it more suitable for its adoption in clinical practice.

Through an automated data management pipeline, the best combination of ensemble tree algorithm and the number of features selected for the model can be inferred. Moreover, in order to detect the best-balanced model in terms of accuracy and explainability, different evaluations are carried out by applying classification and explainability metrics. Therefore, the explainable prediction model implements an Extra Trees classifier over the following 5 features: 'age' (patient's age), 'diabetes' (if the patient has diabetes), 'ejection_fraction' (percentage of blood leaving the heart at each contraction), 'serum_creatinine' (level of creatinine in the blood), and 'time' (follow-up period). The novelty presented by this work is the explainability approach adopted in the prediction models for HF survival which revolves around giving healthcare professionals an easier understanding and interpretability of the outcomes generated by the model. Thus, not only would clinicians be able to early discover a change in a patient's health with a reduced group of indicators, but they could also focus on treating those relevant features to revert a possible situation that might lead to patient death.

As future works, it could be proposed to test the prediction model developed in a clinical setting to test the accuracy robustness of the model with new patients' data as well as to gather insights of healthcare professionals about the explainability of the results provided.

**Funding**

This research did not receive any specific grant from funding agencies in the public, commercial, or not-for-profit sectors.